\documentclass[11pt,letterpaper]{article}
\usepackage{emnlp2017}
\usepackage{times}
\usepackage{latexsym}
\usepackage{graphicx}

\emnlpfinalcopy

\def\emnlppaperid{***}

\title{High-risk learning:\\ acquiring new word vectors from tiny data}

\author{
Aur\'elie Herbelot \\Dept. of Translation and Language Sciences \\Universitat Pompeu Fabra\\{\tt aurelie.herbelot@cantab.net}
\And 
Marco Baroni\\Center for Mind/Brain Sciences\\University of Trento\\{\tt marco.baroni@unitn.it} \\
  }

\date{}

\begin{document}

\maketitle

\begin{abstract}
  Distributional semantics models are known to struggle with small data. It is generally accepted that in order to learn `a good vector' for a word, a model must have sufficient examples of its usage. 
This contradicts the fact that humans can guess the meaning of a word from a few occurrences only. 
In this paper, we show that a neural language model such as Word2Vec only necessitates minor modifications to its standard architecture to learn new terms from tiny data, using background knowledge from a previously learnt semantic space. We test our model on word definitions and on a nonce task involving 2-6 sentences' worth of context, showing a large increase in performance over state-of-the-art models on the definitional task.
\end{abstract}

\section{Introduction}
\label{sec:intro}

Distributional models (DS: \citet{Turney2010,Clark2012,Erk2012}), and in particular neural network approaches \citep{Bengio2003,Collobert2011,Huang2012,Mikolov2013}, do not fare well in the absence of large corpora. That is, for a DS model to learn a word vector, it must have seen that word a sufficient number of times. This is in sharp contrast with the human ability to perform \textit{fast mapping}, i.e. the acquisition of a new concept from a single exposure to information \citep{Lake2011,Trueswell2013,Lake2016}. 

There are at least two reasons for wanting to acquire vectors from very small data. First, some words are simply rare in corpora, but potentially crucial to some applications (consider, for instance, the processing of text containing technical terminology). Second, it seems that fast-mapping should be a prerequisite for any system pretending to cognitive plausibility: an intelligent agent with learning capabilities should be able to make educated guesses about new concepts it encounters.  

One way to deal with data sparsity issues when learning word vectors is to use morphological structure as a way to overcome the lack of primary data  \citep{Lazaridou2013,Luong2013, Kisselew2015,Pado2016}. Whilst such work has shown promising result, it is only applicable when there is transparent morphology to fall back on. Another strand of research has been started by \citet{Lazaridou2016}, who recently showed that by using simple summation over the (previously learnt) contexts of a nonce word, it is possible to obtain good correlation with human judgements in a similarity task. It is important to note that both these strategies assume that rare words are special cases of the distributional semantics apparatus, and thus require separate approaches to model them.

Having different algorithms for modelling the same phenomenon means however that we need some meta-theory to know when to apply one or the other: it is for instance unclear at which frequency a rare word is not rare anymore. Further, methods like summation are naturally self-limiting: they create frustratingly strong baselines but are too simplistic to be extended and improved in any meaningful way. In this paper, our underlying assumption is thus that it would be desirable to build a single, all-purpose architecture to learn word representations from {\it any} amount of data. The work we present views fast-mapping as a component of an incremental architecture: the rare word case is simply the first part of the concept learning process, \textit{regardless of how many times it will eventually be encountered.}

With the aim of producing such an incremental system, we demonstrate that the general architecture of neural language models like Word2Vec \citep{Mikolov2013} is actually suited to modelling words from a few occurrences only, providing minor adjustments are made to the model itself and its parameters. Our main conclusion is that the combination of a heightened learning rate and greedy processing results in very reasonable one-shot learning, but that some safeguards must be in place to mitigate the high risks associated with this strategy.

\section{Task description}

We want to simulate the process by which a competent speaker encounters a new word in known contexts. That is, we assume an existing vocabulary (i.e. a previously trained semantic space) which can help the speaker `guess' the meaning of the new word. To evaluate this process, we use two datasets, described below. 

\paragraph{The definitional nonce dataset}

We build a novel dataset based on encyclopedic data, simulating the case where the context of the unknown word is supposedly maximally informative.\footnote{Data available at \url{http://aurelieherbelot.net/resources/}.} We first record all Wikipedia titles containing one word only (e.g. \textit{Albedo, Insulin}). We then extract the first sentence of the Wikipedia page corresponding to each target title (e.g. \textit{Insulin is a peptide hormone produced by beta cells in the pancreas.}), and tokenise that sentence using the Spacy toolkit.\footnote{\url{https://spacy.io/}}  Each occurrence of the target in the sentence is replaced with a slot (\_\_\_). 

From this original dataset, we only retain sentences with enough information (i.e. a length over 10 words), corresponding to targets which are frequent enough in the UkWaC corpus (\citet{Baroni2009}, minimum frequency of 200). The frequency threshold allows us to make sure that we have a high-quality gold vector to compare our learnt representation to. We then randomly sample 1000 sentences, manually checking the data to remove instances that are, in fact, not definitional. We split the data into 700 training and 300 test instances.

On this dataset, we simulate first-time exposure to the nonce word by changing the label of the gold standard vector in the background semantic space, and producing a new, randomly initialised vector for the nonce. So for instance, \textit{insulin} becomes \textit{insulin\_gold}, and a new random embedding is added to the input matrix for \textit{insulin}. This setup allows us to easily measure the similarity of the newly learnt vector, obtained from one definition, to the vector produced by exposure to the whole Wikipedia. To measure the relative performance of various setups, we calculate the Reciprocal Rank (RR) of the gold vector in the list of all nearest neighbours to the learnt representation. We average RRs over the number of instances in the dataset, thus obtaining a single MRR figure (Mean Reciprocal Rank).

\paragraph{The Chimera dataset}

\begin{figure*}[tb]
\begin{scriptsize}
\begin{verbatim}
Sentences:
Canned sardines and VALTUOR between two slices of wholemeal bread and thinly spread Flora Original.  
@@ Erm, VALTUOR, low fat dairy products, incidents of heart disease for those who have an olive oil rich diet.

Probes: rhubarb, onion, pear, strawberry, limousine, cushion 
Human responses: 3, 2.86, 1.43, 2.14, 1.29, 1.71
\end{verbatim}
\end{scriptsize}
\caption{An example chimera (VALTUOR).}
\label{fig:chimera}
\end{figure*}

Our second dataset is the `Chimera' dataset of \cite{Lazaridou2016}.\footnote{Available at  \url{http://clic.cimec.unitn.it/Files/PublicData/chimeras.zip}.} This dataset was specifically constructed to simulate a nonce situation where a speaker encounters a word for the first time in naturally-occurring (and not necessarily informative) sentences. Each instance in the data is a nonce, associated with 2-6 sentences showing the word in context. The novel concept is created as a `chimera', i.e. a mixture of two existing and somewhat related concepts (e.g., a buffalo crossed with an elephant). The sentences associated with the nonce are utterances containing one of the components of the chimera, randomly extracted from a large corpus.

The dataset was annotated by humans in terms of the similarity of the nonce to other, randomly selected concepts. Fig. \ref{fig:chimera} gives an example of a data point with 2 sentences of context, with the nonce capitalised (\textit{VALTUOR}, a combination of \textit{cucumber} and \textit{celery}). The sentences are followed by the `probes' of the trial, i.e. the concepts that the nonce must be compared to. Finally, human similarity responses are given for each probe with respect to the nonce. Each chimera was rated by an average of 143 subjects. In our experiments, we simply replace all occurrences of the original nonce  with a slot (\_\_\_) and learn a representation for that slot. For each setting (2, 4 and 6 sentences), we randomly split the 330 instances in the data into 220 for training and 110 for testing.

Following the authors of the dataset, we evaluate by calculating the correlation between system and human judgements. For each trial, we calculate Spearman correlation ($\rho$) between the similarities given by the system to each nonce-probe pair, and the human responses. The overall result is the average Spearman across all trials.

\section{Baseline models}
\label{sec:baselines}

We test two state-of-the art systems: a) Word2Vec (W2V) in its Gensim\footnote{Available at \url{https://github.com/RaRe-Technologies/gensim}.} implementation, allowing for update of a prior semantic space; 
b) the additive model of \citet{Lazaridou2016}, using a background space from W2V. 

We note that both models allow for some sort of incrementality. W2V processes input one context at a time (or several, if mini-batches are implemented), performing gradient descent after each new input. The network's weights in the input, which correspond to the created word vectors, can be inspected at any time.\footnote{Technically speaking, standard W2V is not fully incremental, as it requires a first pass through the corpus to compute a vocabulary, with associated frequencies. As we show in \S\ref{sec:experiments}, it however allows for an incremental interpretation, given minor modifications.} As for addition, it also affords the ability to stop and restart training at any time: a typical implementation of this behaviour can be found in distributional semantics models based on random indexing (see e.g. \citealp{QasemiZadeh2017}). This is in contrast with so-called `count-based' models calculated by computing a frequency matrix over a fixed corpus, which is then globally modified through a transformation such as Pointwise Mutual Information.

\paragraph{Word2Vec} We consider W2V's `skip-gram' model, which learns word vectors by predicting the context words of a particular target. The W2V architecture includes several important parameters, which we briefly describe  below.

In W2V, predicting a word implies the ability to distinguish it from so-called \textit{negative samples}, i.e. other words which are \textit{not} the observed item. The number of negative samples to be considered can be tuned. What counts as a context for a particular target depends on the \textit{window size} around that target. W2V features random resizing of the window, which has been shown to increase the model's performance.  Further, each sentence passed to the model undergoes \textit{subsampling}, a random process by which some words are dropped out of the input as a function of their overall frequency. Finally, the \textit{learning rate $\alpha$}  measures how quickly the system learns at each training iteration. Traditionally, $\alpha$ is set low ($0.025$ for Gensim) in order not to overshoot the system's error minimum.

Gensim has an update function which allows us to save a W2V model and continue learning from new data: this lets us simulate prior acquisition of a background vocabulary and new learning from a nonce's context. As background vocabulary, we use a semantic space trained on a Wikipedia snapshot of $1.6B$ words with Gensim's standard parameters (initial learning rate of $0.025$, $5$ negative samples, a window of $\pm 5$ words, subsampling $1e^{-3}$, $5$ epochs). We use the skip-gram model with a minimum word count of $50$ and vector dimensionality $400$. This results in a space with $259,376$ word vectors. We verify the quality of this space by calculating correlation with the similarity ratings in the MEN dataset \citep{Bruni2014}. We obtain $\rho=0.75$, indicating an excellent fit with human judgements.

\paragraph{Additive model} \citet{Lazaridou2016} use a simple additive model, which sums the vectors of the context words of the nonce, taking as context the entire sentence where the target occurs. Their model operates on multimodal vectors, built over both text and images. In the present work, however, we use the semantic space described above, built on Wikipedia text only. We do not normalise vectors before summing, as we found that the system's performance was better than with normalisation. We also discard function words when summing, using a stopword list. We found that this step affects results very positively.

The results for our state-of-the-art models are shown in the top sections of Tables \ref{tbl:definitions} and \ref{tbl:chimeras}. W2V is run with the standard Gensim parameters, under the skip-gram model. It is clear from the results that W2V is unable to learn nonces from definitions ($MRR=0.00007$). The additive model, on the other hand, performs well: an $MRR$ of $0.03686$ means that the median rank of the true vector is $861$, out of a challenging $259,376$ neighbours (the size of the vocabulary). On the Chimeras dataset, W2V still performs well under the sum model -- although the difference is not as marked and possibly indicates that this dataset is more difficult (which we would expect, as the sentences are not as informative as in the encyclopedia case).

\section{Nonce2Vec}

Our system, Nonce2Vec (N2V),\footnote{Code available at \url{https://github.com/minimalparts/nonce2vec}.} modifies W2V in the following ways.

\textbf{Initialisation:} since addition gives a good approximation of the nonce word, we initialise our vectors to the sum of all known words in the context sentences (see \S\ref{sec:baselines}). Note that this is not strictly equivalent to the pure sum model, as subsampling takes care of frequent word deletion in this setup (as opposed to a stopword list). In practice, this means that the initialised vectors are of slightly lesser quality than the ones from the sum model.

\textbf{Parameter choice:} we experiment with higher learning rates coupled with larger window sizes. 
That is, the model should take the risk of a) overshooting a minimum error; b) greedily considering irrelevant contexts in order to increase its chance to learn anything. We mitigate these risks through \textit{selective training} and appropriate \textit{parameter decay} (see below).

\textbf{Window resizing:} we suppress the random window resizing step when learning the nonce. This is because we need as much data as possible and accordingly need a large window around the target. Resizing would make us run the risk of ending up with a small window of a few words only, which would be uninformative.

\textbf{Subsampling:} 
With the goal of keeping most of our tiny data, we adopt a subsampling rate that only discards extremely frequent words.

\textbf{Selective training:} we only train the nonce. That is, we only update the weights of the network for the target. This ensures that, despite the high selected learning rate, the previously learnt vectors, associated with the other words in the sentence, will not be radically shifted towards the meaning expressed in that particular sentence.

Whilst the above modifications are appropriate to deal with the first mention of a word, we must ask in what measure they still are applicable when the term is encountered again (see \S\ref{sec:intro}). With a view to cater for incrementality, we introduce a notion of \textbf{parameter decay} in the system. We hypothesise that the initial high-risk strategy, combining high learning rate and greedy processing of the data, should only be used in the very first training steps. Indeed, this strategy drastically moves the initialised vector to what the system assumes is the right neighbourhood of the semantic space. Once this positioning has taken place, the system should refine its guess rather than wildly moving in the space. We thus suggest that the learning rate itself, but also the subsampling value and window size should be returned to more conventional standards as soon as it is desirable. To achieve this, we apply some exponential decay to the learning rate of the nonce, proportional to the number of times the term has been seen: every time $t$ that we train a pair containing the target word, we set $\alpha$ to $\alpha_0 e^{-\lambda t}$, where $\alpha_0$ is our initial learning rate. We also decrease the window size and increase subsampling rate on a per-sentence basis (see \S\ref{sec:experiments}).

\section{Experiments}
\label{sec:experiments}

\begin{table}
\begin{center}
\begin{tabular}{|l|l|l|}
\hline
 & MRR & Median rank\\
\hline 
W2V & 0.00007 & 111012\\
Sum & 0.03686 & 861\\
\hline
N2V & \textbf{0.04907} & \textbf{623}\\
\hline
\end{tabular}
\caption{Results on definitional dataset}
\label{tbl:definitions}
\end{center}
\end{table}

\begin{table}[t]
\begin{center}
\begin{tabular}{|l|l|l|l|}
\hline
 & L2 $\rho$ & L4 $\rho$ & L6 $\rho$\\
\hline 
W2V & 0.1459 & 0.2457 & 0.2498\\
Sum & {\bf 0.3376} & 0.3624 & {\bf 0.4080}\\ 
\hline
N2V & 0.3320 & {\bf 0.3668}  & 0.3890\\
\hline
\end{tabular}
\caption{Results on chimera dataset}
\label{tbl:chimeras}
\end{center}
\end{table}

We first tune N2V's initial parameters on the training part of the definitional dataset. We experiment with a range of values for the learning rate ($[0.5,0.8,1,2,5,10,20]$), window size ($[5,10,15,20]$), the number of negative samples ($[3,5,10]$), the number of epochs ($[1,5]$) and the subsampling rate ($[500,1000,10000]$). Here, given the size of the data, the minimum frequency for a word to be considered is $1$. The best performance is obtained for a window of 15 words, 3 negative samples, a learning rate of $1$, a subsampling rate of $10000$, an exponential decay where $\lambda = \frac{1}{70}$, and one single epoch (that is, the system truly implements fast-mapping). When applied to the test set, N2V shows a dramatic improvement in performance over the simple sum model, reaching $MMR=0.04907$ (median rank $623$).

On the training set of the Chimeras, we further tune the per-sentence decrease in window size and increase in subsampling. For the window size, we experiment with a reduction of $[1...6]$ words on either side of the target, not going under a window of $\pm 3$ words. Further, we adjust each word's subsampling rate by a factor in the range $[1.1,1.2...1.9,2.0]$. Our results confirm that indeed, an appropriate change in those parameters is required: keeping them constant results in decreasing performance as more sentences are introduced. On the training set, we obtain our best performance (averaged over the 2-, 4- and 6-sentences datasets) for a per-sentence window size decrease of $5$ words on either side of the target, and adjusting subsampling by a factor of $1.9$. Table \ref{tbl:chimeras} shows results on the three corresponding test sets using those parameters. Unfortunately, on this dataset, N2V does not improve on addition.

The difference in performance between the definitional and the Chimeras datasets may be explained in two ways. First, the chimera sentences were randomly selected and thus, are not necessarily hugely informative about the nature of the nonce. Second, the most informative sentences are not necessarily at the beginning of the fragment, so the system heightens its learning rate on the wrong data: the risk does not pay off. This suggests that a truly intelligent system should adjust its parameters in a non-monotonic way, to take into account the quality of the information it is processing. This point seems to be an important general requirement for any architecture that claims incrementality: our results indicate very strongly that a notion of \textit{informativeness} must play a role in the learning decisions of the system. This conclusion is in line with work in other domains, e.g. interactive word learning using dialogue, where performance is linked to the ability of the system to measure its own confidence in particular pieces of knowledge and  ask questions with a high information gain \citep{Yu2016}. It also meets with general considerations on language acquisition, which accounts for the ability of young children to learn from limited `primary linguistic data' by restricting explanatory models to those that provide such efficiency \citep{Clark2010computational}.

\section{Conclusion}

We have proposed Nonce2Vec, a Word2Vec-inspired architecture to learn new words from tiny data. It requires a high-risk strategy combining heightened learning rate and greedy processing of the context. The particularly good performance of the system on definitions makes us confident that it is possible to build a unique, unified algorithm for learning word meaning from any amount of data. However, the less impressive performance on naturally-occurring sentences indicates that an ideal system should modulate its learning as a function of the informativeness of a context sentence, that is, take risks `at the right time'. 

As pointed out in the introduction, Nonce2Vec is designed with a view to be an essential component of an incremental concept learning architecture. In order to validate our system as a suitable, generic solution for word learning, we will have to test it on various data sizes, from the type of low- to middle-frequency terms found in e.g. the Rare Words dataset \citep{Luong2013}, to highly frequent words. We would like to systematically evaluate, in particular, how fast the system can gain an understanding of a concept which is fully equivalent to a vector built from big data. We believe that both quality and speed of learning will be strongly influenced by the ability of the algorithm to detect what we called \textit{informative} sentences. Our future work will thus investigate how to capture and measure informativeness.

\section*{Acknowledgments}

We are grateful to Katrin Erk for inspiring conversations about tiny data and fast-mapping, and to Raffaella Bernardi and Sandro Pezzelle for comments on an early draft of this paper. We also thank the anonymous reviewers for their time and valuable comments. We acknowledge ERC 2011 Starting Independent Research Grant No 283554 (COMPOSES). This project has also received funding from the European Union's Horizon 2020 research and innovation programme under the Marie Sk\l odowska-Curie grant agreement No 751250. 
\includegraphics[scale=0.4]{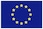}

\bibliography{emnlp2017}
\bibliographystyle{emnlp_natbib}

\end{document}